\documentclass[9pt]{article}

\usepackage{microtype}
\usepackage{subfigure}
\usepackage{booktabs} 

\usepackage{graphicx}
\usepackage{caption}
\usepackage{xspace,color}
\usepackage[dvipsnames]{xcolor}
\usepackage{diagbox,multirow} 
\usepackage{subfigure}
\usepackage{algorithm}
\usepackage{algorithmic}
\usepackage{enumerate}
\usepackage{enumitem,amsmath,amssymb}
\usepackage{hyperref}

\newcommand{\mname}{\texttt{MIMOSA}\xspace }

\usepackage{amssymb,bbm}
\usepackage{amsmath,amsthm,amsfonts,booktabs}
\usepackage[switch]{lineno}

\newtheorem{theorem}{Theorem}

\newtheorem{lemma}{Lemma}
\theoremstyle{definition}
\newtheorem{definition}{Definition}

\newcommand{\RB}{\mathbb{R}}
\newcommand{\calL}{\mathcal{L}}
\newcommand{\calP}{\mathcal{P}}
\newcommand{\calS}{\mathcal{S}}
\newcommand{\calN}{\mathcal{N}}
\newcommand{\calT}{\mathcal{T}}
\newcommand{\haty}{\hat{y}}

\newcommand{\hatz}{\hat{z}}

\newcommand{\f}{\mathbf{f}}
\newcommand{\g}{\mathbf{g}}
\newcommand{\h}{\mathbf{h}}

\begin{document}
%
\title{\mname: Multi-constraint Molecule Sampling for Molecule Optimization\footnote{The paper is accepted by AAAI 2021.}}


\author{Tianfan Fu$^1$ \and Cao Xiao$^2$ \and Xinhao Li$^3$ \and Lucas~M.~Glass$^2$ and Jimeng Sun$^4$}
\date{%
    $^1$Georgia Institute of Technology\\%
    $^2$Analytics Center of Excellence, IQVIA\\
    $^3$North Carolina State University\\
    $^4$University of Illinois at Urbana-Champaign\\ [2ex]%
    August 15, 2020
}
\maketitle

\begin{abstract}
Molecule optimization
is a fundamental task for accelerating drug discovery, with the goal of generating new valid molecules that maximize multiple drug properties while maintaining similarity to the input molecule. Existing generative models and reinforcement learning approaches made initial success, but still face difficulties in simultaneously optimizing multiple drug properties. 
To address such challenges, we propose the MultI-constraint MOlecule SAmpling (\mname) approach, a  sampling framework to use input molecule as an initial guess and sample molecules from the target distribution. 
\mname first pretrains two property-agnostic graph neural networks (GNNs) for molecule topology and substructure-type prediction, where a \textit{substructure} can be either atom or single ring. 
For each iteration, \mname uses the GNNs' prediction and employs three basic substructure operations (\textit{add, replace, delete}) to generate new molecules and associated weights. The weights can encode multiple constraints including similarity and drug property constraints, upon which we select promising molecules for next iteration. 
\mname  enables flexible  encoding of multiple property- and similarity-constraints and can efficiently generate new molecules that satisfy various property constraints and achieved up to 49.1\% relative improvement over the best baseline in terms of success rate. 
The code repository (including readme file, data preprocessing and model construction, evaluation) is available\footnote{\url{https://github.com/futianfan/MIMOSA}}. 
\end{abstract}

\section{Introduction}
\label{sec:intro}

Designing molecules with desirable properties is a fundamental task in drug discovery.
Traditional methods such as high throughput screening (HTS)
tests large compound libraries to identify molecules with desirable properties, which are inefficient and costly~\cite{polishchuk2013estimation,zhavoronkov2018artificial,huang2020moldesigner,lu2019integrated,lu2021cot}. 
Two important machine learning tasks have been studied in this context: 
\begin{itemize}
    \item {\bf Molecule generation} aims at creating new and diverse molecule graphs with some desirable properties~\cite{jin2018junction,You2018-xh}; 
    \item  {\bf Molecule optimization} takes a more targeted approach to find molecule $Y$ with improved drug properties such as drug-likeness and biological activity given an input molecule $X$~\cite{jin2019learning,zhou2019optimization,fu2021probabilistic,lu2022cot}. 
\end{itemize}

Existing works on molecule optimization and molecule generation tasks can be categorized as generative models~\cite{kusner2017grammar,dai2018syntax,gomez2018automatic} and reinforcement learning (RL) methods~\cite{You2018-xh,zhou2019optimization}.
Most existing works only optimize a single property, while multiple properties need to be optimized in order to develop viable drug candidates. 
Recently, \cite{jin2020composing} proposed a molecule generation algorithm that can optimize multiple properties which is a related but different task than molecule optimization since they do not take any specific input molecule as the anchor.     
\cite{Nigam2020Augmenting} proposed a genetic algorithm (GA) for molecule generation and optimization. 
 In this work, we propose a {\it sampling-based strategy} to tackle the {\bf molecule optimization for multi-properties}. 

To allow for flexible and efficient molecule optimization on multiple properties, we propose a new sampling-based molecule optimization framework named MultI-constraint MOlecule SAmpling (\mname). \mname uses the input molecule as an initial guess and pretrains two graph neural networks (GNNs) on molecule topology and substructure-type predictions to produce better molecule embedding for sampling, where \textit{substructure} can be either an atom or a ring.
In each iteration, \mname uses the prediction and employs three basic substructure operations (add, replace, delete) to generate new molecule candidates and associated weights. 
The weights thus effectively encode multiple constraints including similarity to the input molecule and various drug properties, upon which we accept promising molecules for next iteration sampling. 
\mname iteratively produces new molecule candidates and can efficiently draw molecules that satisfy all constraints.
The main contributions of our paper are listed below.
\begin{itemize}
\item \textbf{A new sampling framework for flexible encoding of multiple constraints}. We reformulate molecule optimization task in a sampling framework to draw molecules from the target distribution (Eq.~\eqref{eqn:target_distribution}). The framework provides flexible and efficient encoding of multi-property and similarity constraints as a target distribution (Section~\ref{sec:formulation}). 
\item \textbf{Efficient sampling augmented by GNN pretraining}. With the help of two pretrained GNN models, we designed a Markov Chain Monte Carlo (MCMC) based molecule sampling method that enables efficient sampling from a target distribution (Section~\ref{sec:sampling}). This enables \mname  to leverage vast amounts of molecule data in an unsupervised manner without the need for any knowledge of molecule pairs (i.e., an input molecule and an enhanced molecule) as many existing methods do. 
\item \textbf{Guaranteed unbiased sampling}. We provide theoretical analysis to show that the proposed MCMC method draws unbiased samples from the target distribution, i.e., exhibiting ergodicity and convergence (Section~\ref{sec:theory}).
\end{itemize}
We compare \mname with state-of-the-art baselines on optimizing several important properties across multiple settings, \mname achieves 43.7\% success rate (49.1\% relative improvement over the best baseline GA~\cite{Nigam2020Augmenting}) when optimizing DRD and PLogP jointly.  
\vspace{-0.0cm}

\section{Related Work}
\label{sec:related_work}

\noindent
\textbf{Generative models for molecule optimization}
 project an input molecule to a latent space, then search in the latent space for new and better molecules. For example, \cite{gomez2018automatic}, \cite{Blaschke} utilized SMILES strings as molecule representations to generate molecules. Since string-based approaches often create many invalid molecules, \cite{kusner2017grammar} and \cite{dai2018syntax} designed grammar constraints to improve the chemical validity. 
Recently, \cite{Nigam2020Augmenting} proposed to explore molecule generation using a genetic algorithm. 
 Another line of works focuses on graph representations of molecules, e.g., MolGAN \cite{cao2018molgan}, CGVAE (Constrained Graph VAE)~\cite{liu2018constrained}, JTVAE (Junction Tree VAE) based approaches~\cite{jin2018junction,jin2019learning,jin2019multi}.
Although almost perfect on generating valid molecules, most of them rely on paired data as training data. 

\noindent
\textbf{Reinforcement learning for molecule optimization} 
 are also developed on top of molecule generators for achieving desirable properties. For example, \cite{Olivecrona2017-ry,Putin,fu2022reinforced,popova2018deep,fu2022antibody} applied RL techniques on top of a string generator to generate  SMILES strings. They struggled with the validity of the generated chemical structures. Recently, \cite{You2018-xh}, \cite{zhou2019optimization} leverage deep reinforcement learning to generate molecular graph, achieving perfect validity. However, all these methods require pre-training on a specific dataset, which makes their exploration ability limited by the biases present in the training data. More recently, \cite{jin2020composing} focused on the molecule generation method for creating molecules with multiple properties. However, this approach can lead to arbitrary diverse structures (not optimized for a specific input molecule) and assumes each property is associated with specific molecular substructures that are not applicable to all properties.

In this paper, we proposed a new molecule optimization method that casts molecule optimization as a sampling problem, which provides an efficient and flexible framework for optimizing multiple constraints (e.g., similarity constraint, multiple property constraints) simultaneously.

\section{The \mname Method}
\label{sec:method}


\subsection{Molecule Optimization via Sampling}
\label{sec:formulation}

Slightly different from general molecule generation that focuses on generating valid and diverse molecules, the \textit{molecule optimization} task takes a molecule $X$ as input and aims to obtain a new molecule $Y$ that is not only similar to $X$ but also have more desirable drug properties than $X$. 

We formulate a Markov Chain Monte Carlo (MCMC)- based sampling strategy. MCMC methods are popular Bayesian sampling approaches for estimating posterior distributions and quantifying uncertainty~\cite{liu2001monte,fu2024ddn3,lu2024uncertainty,chen2024uncertainty}. 
They allow drawing samples from complex distributions with desirable sampling efficiency~\cite{welling2011bayesian,chen2021data,wu2022cosbin,zhang2021ddn2} as long as unnormalized probability density for samples can be calculated.

Here to formulate molecule optimization that aims to optimize on the similarity between the input molecule $X$ and the target molecules $Y$ as well as $M$ molecular properties of $Y$,  $\calP_1,\cdots, \calP_M$ (the higher score, the better). We propose to draw $Y$  from the \textit{unnormalized target distribution} in Eq.~\eqref{eqn:target_distribution}.
\begin{equation}
\label{eqn:target_distribution}
\begin{aligned}
p_{X}(Y) 
\propto\ & \mathbbm{1}(Y)\exp\bigg(\eta_0 \text{sim}(X,Y) 
+\eta_1  \Big(\calP_1(Y)-\calP_1(X)\Big) 
\\ & 
+\cdots+ \eta_{M} \Big(\calP_{M}(Y) - \calP_{M}(X)\Big) \bigg)
\end{aligned}
\end{equation}
where $\eta_0, \eta_1, \cdots, \eta_M \in \RB_{+}$ are the hyperparameters that control the strength of various terms, 
$\mathbbm{1}(Y)$ is an indicator function measuring whether the molecule $Y$ is a valid molecule. It is added to ensure the validity of the generated molecule $Y$. 
The target distribution is designed to encode any number of type of constraints, including similarity constraint and multiple drug property constraints. Here the use of $\exp$ is to guarantee $p_X(Y)$ is valid probability distribution. Usually we define the similarity $\text{sim}(X,Y)$ as in Def.~\ref{def:tanimoto} and measured using Eq.~\eqref{eqn:tanimoto}. 

\begin{definition}[\textbf{Tanimoto Similarity of Molecules}]
Denote $\calS_X$ and $\calS_Y$ as fragment descriptor\footnote{Fragment descriptors, represent selected substructures (fragments) of 2D molecular graphs and their occurrences in molecules; they constitute one of the most important types of
molecular descriptors~\cite{baskin2009fragment}.} sets of molecule $X$ and $Y$, respectively. The
Tanimoto similarity between  $X$ and $Y$ is given by 
\begin{equation}
\label{eqn:tanimoto}
\text{sim}(X,Y) = \frac{\vert\calS_X \cap \calS_Y \vert}{\vert \calS_X \cup \calS_Y \vert} \in [0,1],
\end{equation} 
where $\cap, \cup$ represent the intersection and union of two binary vectors respectively; $\vert\cdot \vert$ denotes the cardinality of a set. 
 Higher value means more similar~\cite{wang2024twin,lu2023deep}. 
\label{def:tanimoto}
\end{definition}
\vspace{-0.0cm}

\begin{table}[tb]
\small 
\centering
\caption{ Notations used in the paper. }
\resizebox{0.8\columnwidth}{!}{
\begin{tabular}{c|l}
\toprule[1pt]
Notations & short explanation \\ 
\hline 
$X$, $Y$ & Input molecule, target molecule. \\ 
$\text{sim}(X,Y) \in [0,1]$ & Similarity of molecules $X$ and $Y$. \\
$p_{X}(Y) $ & Target dist. when optimizing $X$, Eq.~\ref{eqn:target_distribution}.  \\
$M$ & \# of properties to optimize. \\
$\gamma_0, \gamma_1, \cdots, \gamma_M \in \RB_{+}$ & Hyperparameter in Target dist. $p_{X}(Y)$. \\ 
$\calP_1,\cdots, \calP_M$ & Molecular properties to optimize.  \\ 
$\mathbbm{1}(Y)$ & Validity Indicator func. of molecule $Y$.  \\
$K$ & Depth of GNN. \\ 
$\h_v^{(k)} \in \RB^{300}$ & Node embedding $v$ in the $k$-th layer. \\
$C_1/C_2$ & \# of all possible substructures/bonds. \\
$v; s_v/s'_{v}$ & node $v$; substructures of $v$. \\
$\f_v/\g_e$ & one-hot node/edge feature. \\
$\mathbf{\haty}_v/$mGNN($Y,v$) & substructure distribution. Eq.~\eqref{eqn:mask_prediction}. \\ 
$\hatz_v$/bGNN($Y,v$) & probability of $v$ will expand. Eq.~\eqref{eqn:bGNN}. \\
$\mathbf{y}_v/z_v$ & ground truth label of node $v$\\
$Y/Y'$ & current/next Sample. \\ 
$S_{\text{add}}, S_{\text{replace}}, S_{\text{delete}} $ & sampling operation from $Y$ to $Y'$. \\
\bottomrule[1pt]
\end{tabular}}
\label{table:notation}
\end{table}

\subsection{The \mname Method for Molecule Sampling}
\label{sec:sampling}
\begin{figure*}[htb]
\centering
\includegraphics[width=\textwidth]{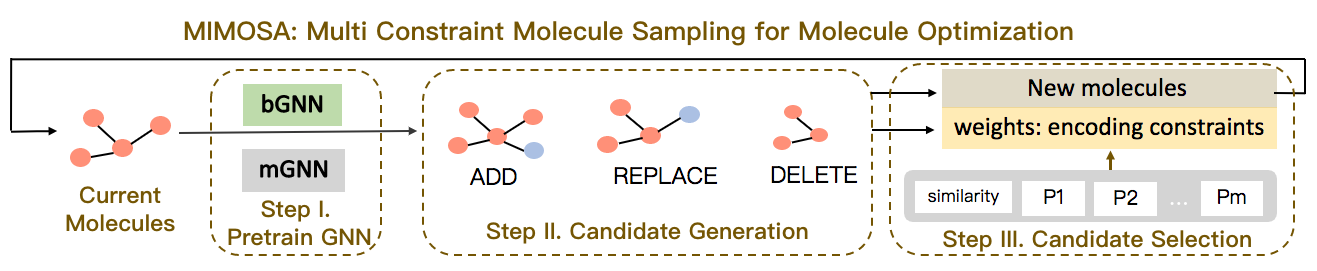}
\caption{The Multi-constraint Molecule Sampling for Molecule Optimization(\mname)  framework illustrated using a single  molecule. In Step I (Pretrain GNN), \mname pretrains two property-agnostic GNNs for molecule topology and substructure-type prediction. Then, in Step II (Candidate Generation), \mname uses the prediction and employs three basic substructure operations (ADD, REPLACE and DELETE) to generate new molecule candidates. 
In Step III (Candidate Selection), \mname assigns weights for new molecule. The weights can encode multiple constraints including similarity and drug property constraints, upon which we accept promising molecules for next iteration. 
\mname iteratively edits the molecule and can efficiently draw molecule samples.}
\label{fig:substructure_edit}
\vskip -10pt
\end{figure*}

Fig.~\ref{fig:substructure_edit} illustrates the overall procedure of \mname, which can be decomposed into the following steps: (1) \underline{Pretrain GNN}. 
\mname pre-trains two graph neural networks (GNNs) using many unlabeled molecules, which will be used in the sampling process. Then \mname iterates over the following two steps. (2) \underline{Candidate Generation}. We generate and score molecule candidates via modification operations ({\it add, delete, replace}) to the current molecule. (3) \underline{Candidate Selection}. We perform MCMC sampling to select promising molecule candidates for the next sampling iteration by repeating Step 2 and 3. Note that all modification operations are on the \textit{substructure} level, where a substructure can be either an atom or a single ring. The substructure set includes all 118 atoms and 31 single rings. 



\subsection*{(I) Pretrain GNNs for  Substructure-type and Molecule Topology Prediction}

To provide accurate molecule representation, we propose to pretrain molecule embeddings on large molecule datasets. 
Since we consider molecules in graph representations where each substructure is a node, we develop two GNN based pretraining tasks to assist molecule modification. These two GNNs will assess the probability of each substructure conditioned on all the other substructures in the molecule graph.

Mathematically, in molecular graph $Y = (V,E)$, we have one-hot node feature $\f_v \in \{0,1\}^{C_1}$ for every node $v \in V$ and one-hot edge feature $\g_{e} \in  \{0,1\}^{C_2}$ for every edge $e=(u,v)\in E$.  $C_1$ and $C_2$ are the number of substructures and the number of bond types, respectively. In our experiment, $C_1 = 149$, including 118 atoms, 31 single rings, and $C_2 = 4$ correspond to the four bond types. 
We list the node and edge features in the appendix.

The two  Graph Neural Networks (GNN) ~\cite{hu2019strategies} are learned with these node and edge features and the same molecule graph to learn an embedding vector $\h_v$ for every node $v \in V$. 
\begin{equation} 
\label{eqn:gnn_output}
\begin{aligned}
\h_{v}^{(k)} = & \mathrm{ReLU}\bigg( \mathrm{MLP} \Big(\mathrm{CONCAT} 
\\ & 
\ \big(\sum_{u \in \calN(v) \cup \{v\}} \h_{u}^{(k-1)}, \sum_{e=(u,v): u\in \calN(v)} \g_{e}^{(k-1)} \big)\Big) \bigg),  \ \  
\end{aligned}
\end{equation}
where the layer $k=1,\cdots,K$; $\text{CONCAT}(\cdot,\cdot)$ is the concatenation of two vectors; $\calN(v)$ is the set of all neighbors of $v$; $\h_v^{(0)}$ is the initial node embedding $\f_v$. 
After $K$ layers of GNN, we have the final node embedding $\h_v^{(K)}$ for node $v$. In our experiment, $K = 5$. 

Using the same GNN architecture, we trained  two GNN models: one for substructure-type prediction called  {\it mGNN} and one for molecule topology prediction called {\it bGNN}:
We choose to train two separate GNNs instead of sharing a single GNN because sufficient unlabeled molecule samples exist and the two tasks are very different in nature.

The \textbf{mGNN} model aims at multi-class classification for predicting the substructure type of a masked node. The mGNN model outputs the type of an individual substructure conditioned on all other substructures and their connections. We mask the individual substructure and replace it with a special masked indicator following~\cite{hu2019strategies}.
Suppose we only mask one substructure for each molecule during training and $v$ is the masked substructure (i.e., node), $y_v$ is the node label corresponding to masked substructure type, we add fully-connected (FC) layers with softmax activation (Eq.~\eqref{eqn:softmax}) to predict the type of the node $v$.
\begin{equation}
\label{eqn:softmax}
\begin{aligned}
\mathbf{\haty}_{v} = \mathrm{Softmax}\big(\mathrm{FC}(\h_v^{(K)})\big). 
\end{aligned}
\end{equation}
where $\mathbf{\haty_v}$ is a $C_1$ dimension vector, indicating the predicted probability of all possible substructures. 
Multi-class cross entropy loss (Eq.~\eqref{eqn:mGNN}) is used to guide the training of GNN: 
\begin{equation}
\label{eqn:mGNN}
\begin{aligned}
\calL(\mathbf{y}_v, \mathbf{\haty}_v) = - \sum_{i=1}^{C_1} \big( (\mathbf{y}_v)_{i} \log (\mathbf{\haty}_v)_{i} \big),
\end{aligned}
\end{equation}
where $\mathbf{y}_v$ is the groundtruth, one-hot vector. $C_1$ is number of all substructures (atoms and single rings), $(\mathbf{y}_v)_{i}$ is i-th element of vector $\mathbf{y}_v$.

To summarize, the prediction of mGNN is defined as 
\begin{equation}
\label{eqn:mask_prediction}
\mathbf{\haty}_{v} \triangleq \text{mGNN}(Y, \text{mask} = v) = \text{mGNN}(Y, v),
\end{equation}
where in a given molecule $Y$ the node $v$ is masked, mGNN predicts the substructure distribution on masked node $v$, which is denoted $\mathbf{\haty}_{v}$.

The \textbf{bGNN} model aims at binary classification for predicting the molecule topology.
The goal of bGNN is to predict whether a node will expand. To provide training labels for bGNN, we set the leaf nodes (nodes with degree 1) with label $z_v=0$ as we assume they are no longer expanding. And we set label $z_v=1$ on the non-leaf nodes that are adjacent to leaf nodes as those nodes expanded (to the leaf nodes). 
The prediction is done via
\begin{equation}
\label{eqn:bGNN2}
\begin{aligned}
\hatz_{v} = \mathrm{Sigmoid}\big(\mathrm{FC}(\h_v^{(K)}) \big),
\end{aligned}
\end{equation}
where FC is two-layer fully-connected layers (of 50 neurons followed by 1 neuron). $\h_v^{(K)}$ is defined in Eq.~\eqref{eqn:gnn_output}, the node embedding of $v$ produced by GNN. 
Binary cross-entropy loss is used to guide the training:
\begin{equation}
\label{eqn:bGNN1}
\begin{aligned}
\calL(z_v, \hatz_v) = - z_v\log(\hatz_v) - (1 - z_v)\log (1 - \hatz_v). 
\end{aligned}
\end{equation}
Since the total number of unlabeled molecules is large, when training bGNN we randomly select one substructure $v$ for each molecule to speed up the pretraining. 

In sum, prediction of bGNN is defined as 
\begin{equation}
\label{eqn:bGNN}
\hatz_v \triangleq \text{bGNN}(Y, v), 
\end{equation}
where v is a node in molecule $Y$, $\hatz_v$ is the probability that $v$ will expand. 
\vspace{-0.0cm}

\subsection*{(II) Candidate Generation via Substructure Modification Operation}
With the help of mGNN and bGNN, we define substructure modification operations namely {\it  replace, add or delete} on input molecule $Y$:

\begin{itemize}[leftmargin=*]
\item \underline{Replace a substructure}. 
At node $v$, the original substructure category is $s_v$.
\begin{enumerate}
\item We mask $v$ in $Y$, evaluate the substructure distribution in $v$ via mGNN, i.e., $\mathbf{\haty}_{v} = \text{mGNN}(Y, v),$ as Eq.~\eqref{eqn:mask_prediction}. 
\item Then we sample a new substructure $s'_v$ from the multinomial distribution $\mathbf{\haty}_v$, denoted by $s'_v \sim \text{Multinomial}(\mathbf{\haty}_v)$.    
\item At node $v$, we replace the original substructure $s_v$ with new substructure $s'_v$ to produce the new molecule $Y'$. 
\end{enumerate}
The whole operation is denoted as 
\begin{equation}
\label{eqn:replace}
Y' \sim S_{\mathrm{replace}}(Y'|Y).
\end{equation}

\item \underline{Add a substructure}. 
Suppose we want to add a substructure as leaf node (denoted as $v$) connecting to an existing node $u$ in current molecule $Y$. The substructure category of $v$ is denoted $s_v$, which we want to predict. 
\begin{enumerate}
 \item We evaluate the probability that node $u$ has a leaf node $v$ with help of bGNN in Eq.~\eqref{eqn:bGNN}, i.e., 
\begin{equation*}
  \hatz_u = \mathrm{bGNN}(Y, u) \in [0, 1].   
\end{equation*}
 \item Suppose the above prediction is to add a leaf node $v$. We then generate a new molecule $Y'$ via adding $v$ to $Y$ via a new edge $(u,v)$.   
 \item In $Y'$, $s_v$, the substructure of $v$ is unknown. We will predict its substructure  using mGNN, i.e., $\mathbf{\haty}_{v} = \text{mGNN}(Y', v)$, following Eq.~\eqref{eqn:mask_prediction}. 
 \item We sample a new substructure $s'_v$ from the multinomial distribution $\mathbf{\haty}_v$ and complete the new molecule $Y'$. 
\end{enumerate}
The whole operation is denoted as 
\begin{equation}
\label{eqn:add}
Y' \sim S_{\mathrm{add}}(Y'|Y).
\end{equation}

\item \underline{Delete a substructure}. 
We delete a leaf node  $v$ in current molecule $Y$. 
It is denoted 
\begin{equation}
\label{eqn:delete}
Y' \sim S_{\mathrm{delete}}(Y'|Y).
\end{equation}
\end{itemize}

In the MCMC process, $S_*(Y'|Y)$ indicates the sequential sampling process from previous sample $Y$ to next sample $Y'$. And the very first sample is the input $X$. 

\noindent\underline{Handling Bond Types and Rings}. 
Since the number of possible bonds are small (single, double, triple, aromatic), we enumerate all and choose the one with largest $p_X(Y)$. 
In some case, basic operation would generate invalid molecules. Based on the indicator function in target distribution in Eq.~\eqref{eqn:target_distribution}, the density is equal to 0. Thus, we perform validity check using RDKit~\cite{landrum2006rdkit} to filter out the new molecule graphs that are not valid. 
When adding/replacing a ring, there might be multiple choices to connect to its neighbor. We enumerate all possible choices and retain all valid molecules.  

\vspace{-0.0cm}

\subsection*{(III) Candidate Selection via  MCMC Sampling}

The set of generated candidate molecules can be grouped as three sets based on the type of substructure modification they received, namely, {\it replace} set $S_{\mathrm{replace}}$, {\it add} set $S_{\mathrm{add}}$, and {\it delete} set $S_{\mathrm{delete}}$. \mname uses the  Gibbs sampling~\cite{geman1984stochastic}, a particular type of MCMC, for molecule candidate selection.  Gibbs sampling algorithm generates an instance from the distribution of each variable in sequential or random order~\cite{levine2006optimizing}, conditional on the current values of the other variables. Here molecules from the three sets will be sampled with different sampling weights. Their  weights are designed to satisfy the detailed balance condition~\cite{brooks2011handbook}. \\

\noindent
\underline{Sampling $S_{\mathrm{replace}}$}.
For molecules produced by the ``replace'' operation, the weight in sampling $w_r$ is given by Eq.~\eqref{eqn:replace_AC}.
\begin{equation}
\label{eqn:replace_AC}
 w_r = \frac{ p_{X}(Y')\cdot [\text{mGNN}(Y,v)]_{s'_v} }{ p_{X}(Y)\cdot [\text{mGNN}(Y,v)]_{s_v} },
\end{equation}
where $P_{X}(\cdot)$ is the unnormalized target distribution for optimizing $X$, defined in Eq.~\eqref{eqn:target_distribution}, $[\text{mGNN}(Y,v)]_{s_v}$ is the predicted probability of the substructure $s_v$ in the prediction distribution $\text{mGNN}(Y, v)$. 
The acceptance rate in the proposal is $\min\{1, w_r\}$.  
If the proposal is accepted, we use the new prediction $s'_v$ to replace origin substructure $s_v$ in current molecule $Y$ and produce the new molecule $Y'$.  \\

\noindent
\underline{Sampling $S_{\mathrm{add}}$}. For molecules produced by the ``add'' operation, the weight in sampling is given by Eq.~\eqref{eqn:add_AC}.
\begin{equation}
\label{eqn:add_AC}
w_a = \frac{p_X(Y') \cdot \text{bGNN}(Y, u) \cdot [\text{mGNN}(Y',v)]_{s_v}  }{ p_X(Y) \cdot (1 - \text{bGNN}(Y, u)) },
\end{equation}
where 
The  acceptance rate in the proposal is $\min\{1, w_a\}$.

\noindent
\underline{Sampling $S_{\mathrm{delete}}$}. For these molecules produced by ``delete'' operation, the weight in sampling is given by Eq.~\eqref{eqn:delete_AC}.
\begin{equation}
\label{eqn:delete_AC}
\begin{aligned}
\hspace*{-0.3cm}
w_d = \frac{ p_X(Y') \cdot \big(1 - \text{bGNN}(Y', u) \big) }{ p_X(Y) \cdot \text{bGNN}(Y', u) \cdot [\text{mGNN}(Y,v)]_{s_v} },
\end{aligned}
\end{equation}
where $v$ is the deleted node, leaf node (with degree 1) in molecular graph of $Y$. 
$u$ and $v$ are connected in $Y$. 
The acceptance rate in the proposal is $\min\{1, w_d\}$.

\noindent
\textit{Soft-constraint Encoding}. For these operations, any number or type of constraints (e.g., here the similarity and drug property constraints) can be encoded in $p_X(Y)$ and $p_X(Y')$ and thus reflected in the weights  $w_r, w_a, w_d$. 

For a single-chain MCMC, we construct the transition kernel as given by Eq.~\eqref{eqn:proposal}.
\vspace{-0.1em}
\begin{equation}
\label{eqn:proposal}
\begin{aligned}
Y' \sim \left\{
\begin{aligned}
& S_{\mathrm{replace}}(Y'  |\ Y), &\text{prob}\  \gamma_1,  \text{accept w.}\ \min\{1,w_r\}, \\
& S_{\mathrm{add}}(Y' |\ Y), & \text{prob.}\  \gamma_2,  
 \text{accept w.}\ \min\{1,w_a\},    \\
& S_{\mathrm{delete}}(Y' |\ Y), & \text{prob.}\  \gamma_3,  \text{accept w.}\ \min\{1,w_d\},  \\ 
\end{aligned}
\right.
\end{aligned}
\end{equation}
where  $\gamma_1, \gamma_2, \gamma_3 \in \RB_{+}$ are hyperparameters that determine the sampling probabilities from the three molecule sets. 
In Section~\ref{sec:theory}, we show the transition kernel will leave the target distribution $p_X(Y)$ invariant for arbitrary $\gamma_1,\gamma_2,\gamma_3$ satisfying $\gamma_1 + \gamma_2 + \gamma_3 = 1$ and $\gamma_2=\gamma_3$. 
After molecules are sampled, they will be accepted with their corresponding acceptance probabilities related to $w_r, w_a, w_d$. 

\begin{algorithm}[h!]
\caption{ \mname for Molecule Optimization } \label{alg:main}
\begin{algorithmic}[1]
\STATE \textbf{Input}: molecule $X$, \# of Particle $N$, max \# of sampling iter. $T_\mathrm{max}$, \# of burn-in iter. $T_\mathrm{burnin}$
\STATE \textbf{Output}: Generated molecules $\Phi$. 
\STATE \# Step (I) Pretrain GNN
\STATE Train mGNN (Eq.\ref{eqn:mask_prediction}), bGNN (Eq.\ref{eqn:bGNN}). \ 
\STATE Candidate set $\Theta = \{X\}$, Output set $\Phi = \{\}$. 
\FOR{$\mathrm{iter} = 1,\cdots, T_\text{max}$} 
\STATE \# Step (II) Candidate Generation. 
\STATE Candidate Pool $\Psi = \{\}$. 
\FOR{molecule $Z$ in $\Theta$}
\STATE Generate candidates  $Z'$ via editing $Z$ using substructure operations; validity check; add $Z'$ in $\Psi$.   \ \ \ \  
\ENDFOR
\STATE $\Theta = \{\}$.
\STATE \# Step (III) Candidate Selection. 
\IF{$\mathrm{iter} < T_\mathrm{burnin} $} 
\STATE Select $N$ molecules with highest density value (Eq.~\ref{eqn:target_distribution}) from $\Psi$ and add them into $\Theta$. 
\ELSE 
\STATE Draw $N$ molecules from $\Psi$ using importance sampling ($\propto$ weight $w_r$ in Eq.~\eqref{eqn:replace_AC}, $w_a$ in Eq.~\eqref{eqn:add_AC} or $w_d$ in Eq.~\eqref{eqn:delete_AC}) and add to $\Theta$. 
\ENDIF 
\STATE $\Phi = \Phi \cup \Theta$. 
\ENDFOR 
\end{algorithmic}
\end{algorithm}

The \mname method is summarized in Algorithm~\ref{alg:main}. 
To accelerate the sampling procedure, we also deploy a multi-chain strategy~\cite{liu2000multiple}: during each step, we use $N$ samples for each state, with each sample generating multiple proposals. Also, during burn-in period (Step 12 in Algorithm~\ref{alg:main}), we pick  the molecules with highest density  for efficiency~\cite{brooks2011handbook}. 
We  retain N proposals in iterative sampling.

\vspace{-0.0cm}

\subsection{Analysis of the MCMC Algorithm}
\label{sec:theory}

Our MCMC method  draws unbiased samples from the target distribution, i.e., exhibiting ergodicity and convergence.  We defer the proofs of Lemma~\ref{lemma:ergodic} and ~\ref{lemma:detailed_balance} to the appendix.
\begin{theorem}
\label{thm:main}
Suppose $\{Y_1,Y_2,\cdots, Y_n\}$ is the chain of molecules sampled via MCMC based on transition kernel defined in Eq.~\eqref{eqn:proposal}, with initial state $X$, then the Markov chain is ergodic with stationary distribution $p_X(Y)$ in Eq.~\eqref{eqn:target_distribution}. That is, empirical estimate (time average over $Y_1,Y_2,\cdots,Y_n$) is equal to target value (space average over $p_X(Y)$), i.e., $\lim\limits_{n\xrightarrow{}+\infty} \frac{1}{n} \sum_{i=1}^{n} f(Y_i) = \int f(Y) p_X(Y) dY $
holds for any integratable function $f$. 
\end{theorem}

\noindent\textbf{Proof Sketch}. 
We split the proof into  Lemma~\ref{lemma:ergodic} and ~\ref{lemma:detailed_balance}. 
First, regarding the ergodicity, it is sufficient to prove the irreducibility, aperiodicity of the Markov chain (Lemma~\ref{lemma:ergodic}). 
Then, to show that $p_X(Y)$ is maintained invariant for the whole chain, in Lemma~\ref{lemma:detailed_balance}, following~\cite{brooks2011handbook}, we show that detailed balance condition holds for any neighboring samples ($Y_{i}$ and $Y_{i+1}$).
Then we strengthen this results on the whole chain. 
\begin{lemma}
\label{lemma:ergodic} 
The Markov chain of the sampled molecules ($\{Y_1, \cdots,Y_n\}$, starting at $X$, based on transition kernel  in Eq.~\eqref{eqn:proposal}) is ergodic over the target distribution $p_X(Y)$. 
\end{lemma}

\begin{lemma}
\label{lemma:detailed_balance}
$p_X(Y)$ is maintained as the invariant distribution for the whole Markov chain produced by MCMC transition kernel defined in Eq.~\eqref{eqn:proposal}. 
\end{lemma}

\vspace{-0.0cm}

\section{Experiment}
\label{sec:experiment}

\subsection{Experimental Setup}
\label{sec:setup}

\newcommand{\figurewidth}{2.2}

\textbf{Dataset and Molecular Properties}.  We use 2 million molecules from ZINC database~\cite{sterling2015zinc,hu2019strategies} to train both mGNN and bGNN. 
Following~\cite{You2018-xh,jin2019learning,zhou2019optimization,jin2019multi,fu2021probabilistic,fu2020alpha,fu2021differentiable}, we focus on the molecular properties below. For all scores, the higher the better. 

\begin{itemize}[leftmargin=*]
\item \underline{QED} (Quantitative Estimate of Drug likeness) is an indicator of drug-likeness~\cite{bickerton2012quantifying}. 
\item \underline{DRD} (Dopamine Receptor)  measures a molecule's biological activity against a biological target  dopamine type 2 receptor~\cite{comings1996dopamine}.  
\item \underline{PLogP} (Penalized LogP)  is the log of the partition
ratio of the solute between octanol and water minus the synthetic accessibility score and number of long cycles~\cite{ertl2009estimation}. 
\end{itemize}
Note that PLogP is more sensitive to the change of local molecule structures, while DRD and QED are related to both local and global molecule structures. 
For chemically valid molecules, their QED, DRD2 and LogP scores can be evaluated via Therapeutic Data Common (TDC)~\cite{huang2021therapeutics} (using two lines of code, a python library).

\noindent
\textbf{Baseline Methods}.
We compare \mname with the following molecule optimization baselines. The parameter setting of these methods are provided in the appendix.
\begin{itemize}[leftmargin=*]
 \item \underline{JTVAE} (Junction Tree Variational Auto-Encoder)~\cite{jin2018junction} is a  generative model that learns latent space to generate desired molecule. 
It also uses an encoder-decoder architecture and leverage a junction tree to simplify molecule generation procedure.
 \item \underline{VJTNN} (Variational Junction Tree Encoder-Decoder)~\cite{jin2019learning} improves over JTVAE by leveraging adversarial learning  and attention. 
 \item \underline{GCPN} (Graph Convolutional Policy Network)~\cite{You2018-xh}. GCPN is state-of-the-art reinforcement learning based approach on molecule optimization. It leverages graph convolutional policy networks to generate molecular structures with specific property, where molecular property is included in reward. 
 \item \underline{GA} (Genetic Algorithm)~\cite{Nigam2020Augmenting} is a genetic algorithm that explores chemical space efficiently. 
\end{itemize}
Details on \textbf{Implementation, Features, Dataset Construction, Evaluation Strategies} are in Appendix. 

\noindent\textbf{The code repository} (including readme file, data preprocessing and model construction, evaluation) is available at \underline{\url{https://github.com/futianfan/MIMOSA}}.

\noindent
\textbf{Metrics} We consider the following metrics for evaluation. 
\begin{itemize}[leftmargin=*]
\item \underline{Similarity} between the input and  generated molecule, measured by Tanimoto similarity over Morgan fingerprints~\cite{rogers2010extended}, defined in Eq.~\eqref{eqn:tanimoto}. 
\item \underline{Property Improvement} of generated molecule in QED, DRD, and PLogP. It is defined as the  difference of the property score between generated molecules $Y$ and input molecule $X$, i.e., $\text{property}(Y) - \text{property}(X)$. 
\item \underline{Success Rate (SR)} based on similarity and property improvement between input molecule $X$ and generated molecule $Y$. We follow the same definitions of SR as in \cite{jin2019learning} (See details in appendix).

\end{itemize}



\subsection{Results}

\subsection*{Exp 1. Optimize Multiple Properties}

To evaluate model performance on  optimizing multiple drug properties, we consider the following combinations of property constraints:

\noindent (1) optimize  QED (drug likeness) and PLogP (solubility); 

\noindent (2) optimize  DRD (biological activity against dopamine type 2 receptor ) and PLogP (solubility).

\begin{table}[h!]
\small
\centering
\hspace{-0.1in}
\caption{\textbf{Exp 1}.  Optimizing Multiple Properties. 
}
\label{table:multi}
 \resizebox{0.67\textwidth}{!}{
\begin{tabular}{lcccc}
\toprule[0.6pt]
\multicolumn{5}{c}{\bf Optimizing  PLogP and QED} \\ 
Method & Similarity & PLogP-Imp. & QED-Imp. & Success \\ \hline
JTVAE & 0.16$\pm$0.08 & 0.14$\pm$0.27 & 0.01$\pm$0.10 & 0.4\% \\
VJTNN & 0.17$\pm$0.06 & 0.46$\pm$0.35 & 0.02$\pm$0.09 & 1.0\% \\
GCPN & 0.25$\pm$0.15 & 0.56$\pm$0.25 & 0.06$\pm$0.08 & 11.3\% \\
GA &  0.35$\pm$0.16 & \bf  0.93$\pm$0.67 & 0.09$\pm$0.07 & 24.9\% \\ \hline
\mname & \bf 0.42$\pm$0.17 & \bf 0.93$\pm$0.48 & \bf 0.10$\pm$0.09 & \bf 32.0\%\\
\midrule
 \multicolumn{5}{c}{\bf Optimizing  PLogP and DRD} \\ 
Method &  Similarity & PLogP-Imp. & DRD-Imp. & Success  \\ \hline
JTVAE & 0.18$\pm$0.08 & 0.20$\pm$0.18 & 0.18$\pm$0.09 & 0.8\% \\
VJTNN & 0.18$\pm$0.08 & 0.55$\pm$0.16 & 0.27$\pm$0.05 & 3.4\%\\
GCPN & 0.23$\pm$0.12 & 0.38$\pm$0.25 & 0.25$\pm$0.11 & 20.4\%\\
GA & 0.38$\pm$0.16 & 0.68$\pm$0.49 & 0.20$\pm$0.16 & 29.3\%\\ \hline
\mname & \bf 0.54$\pm$0.16 & \bf 0.75$\pm$0.48 & \bf 0.35$\pm$0.20  & \bf 43.7\% \\
\bottomrule[0.6pt]
\end{tabular}}
\end{table}

\begin{figure*}[h!]
\centering
\subfigure[Input Molecule $X$; QED:0.72; PLogP:-3.60 ]{
\includegraphics[width=\figurewidth cm]{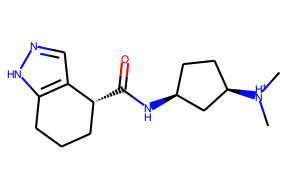}}
\hspace{.1in}
\subfigure[sim: 0.66; QED: 0.93; PLogP: -1.2]{
\includegraphics[width=\figurewidth  cm]{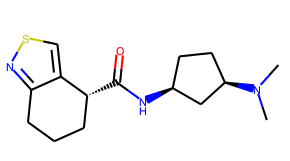}}
\hspace{.1in}
\subfigure[sim: 0.59; QED: 0.93; PLogP: -1.1]{
\includegraphics[width=\figurewidth cm]{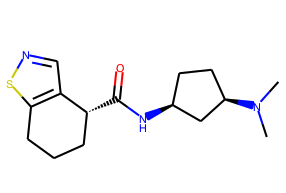}}
\hspace{.1in}
\subfigure[sim: 0.57; QED: 0.92; PLogP:-1.6]{
\includegraphics[width=\figurewidth cm]{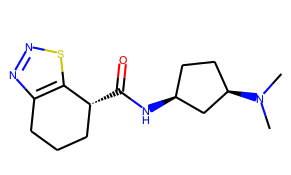}} \\
\subfigure[Input Molecule $X$; QED: 0.71; PLogP: -3.9]{
\includegraphics[width=\figurewidth cm]{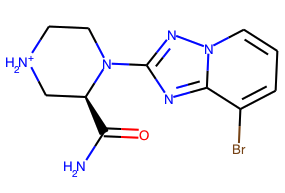}}
\hspace{.1in}
\subfigure[sim: 0.837; QED: 0.90; PLogP: -0.6]{
\includegraphics[width=\figurewidth cm]{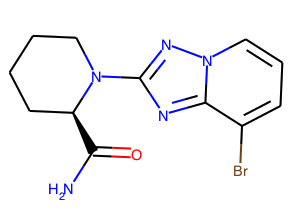}}
\hspace{.1in}
\subfigure[sim: 0.872; QED: 0.89; PLogP: -1.2]{
\includegraphics[width=\figurewidth cm]{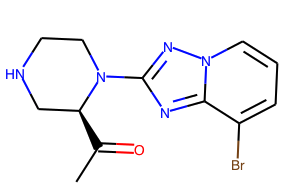}}
\hspace{.1in}
\subfigure[sim: 0.812; QED: 0.88; PLogP: -1.4]{
\includegraphics[width=\figurewidth cm]{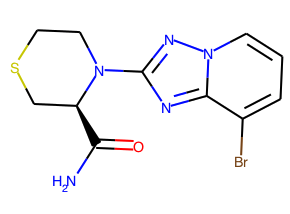}}
\caption{\textbf{Exp 3}. Examples of ``QED \& PLogP'' optimization. (\textbf{Upper}), the imidazole ring in the input molecule (a) is replaced by less polar rings thiazole (b and c) and thiadiazol (d). Since  more polar indicates lower PLogP, the output molecules increase  PLogP while maintaining the molecular scaffold. \textbf{Lower}), the PLogP of input molecule (e) is increased by neutralizing the ionized amine (g) or replacing with substructures with less electronegativity (f and h). These changes  improve the QED.}
\label{fig:example}
\vskip -12pt
\end{figure*}

From Table~\ref{table:multi},   \mname has significantly better and stable performance on all metrics, with $28.5\%$ relative higher success rate in  optimizing both QED and PLogP, and  $49.1\%$ relative higher success rate in  optimizing both DRD and PLogP compared with the second best algorithm GA. The GA algorithm  uses genetic algorithm for local structure editing, hence is expected to work well on optimizing properties that are sensitive to local structural changes, such as joint optimizing both QED  and PLogP where PLogP is related to the polarity of a molecule and is sensitive to the change of local structure. Because of the local editing of GA,  GA does not perform well on optimizing both DRD  and PLogP since DRD is less sensitive to the change of local structures. Among other baselines, GCPN has better performance. However, its performance is not stable when optimizing PLogP and QED simultaneously, since it can generate molecules with negative QED improvement. 

\begin{table}[h!]
\small 
\vskip -2pt
\caption{\textbf{Exp 2}. Optimizing Single Property. 
}
\label{table:single}
\hspace{-0.1in}
\begin{minipage}[b]{\linewidth}
\centering
 \resizebox{0.67\textwidth}{!}{
\begin{tabular}{lccc}
\toprule[0.6pt]
 \multicolumn{4}{c}{\bf Optimizing QED} \\
Method & Similarity & QED-Improve & Success  \\ \hline 
JTVAE & 0.30$\pm$0.09 & 0.17$\pm$0.12 & 17.4\%    \\ 
VJTNN & 0.37$\pm$0.11 & \bf 0.20$\pm$0.05 &  37.6\%   \\ 
GCPN & 0.32$\pm$0.14 & \bf 0.20$\pm$0.09 & 26.5\%     \\ 
GA & 0.43$\pm$0.17 & 0.17$\pm$0.11  & 42.5\%  \\ \hline
\mname & \bf 0.50$\pm$0.30 & \bf0.20$\pm$0.14 &  \bf 47.8\% \\ \midrule
 \multicolumn{4}{c}{\bf Optimizing DRD} \\ 
Method & Similarity & DRD-Improve & Success  \\ \hline 
JTVAE & 0.31$\pm$0.07 & 0.34$\pm$0.17 &  25.6\%   \\ 
VJTNN & 0.36$\pm$0.09  & 0.40$\pm$0.20 &  40.5\%   \\ 
GCPN & 0.30$\pm$0.07 & 0.35$\pm$0.20 & 27.8\%  \\ 
GA & 0.46$\pm$0.14 & 0.25$\pm$0.10 &  37.5\% \\ \hline
\mname & \bf 0.57$\pm$0.29 & \bf 0.43$\pm$0.29 &  \bf 48.3\%  \\ 
\midrule
 \multicolumn{4}{c}{\bf Optimizing PLogP} \\
Method & Similarity & PLogP-Improve & Success  \\ \hline 
JTVAE & 0.30$\pm$0.09 & 0.28$\pm$0.17 & 2.9\%  \\ 
VJTNN & 0.38$\pm$ 0.08 & 0.47$\pm$0.24 & 14.3\%  \\ 
GCPN & 0.32$\pm$ 0.07 & 0.33$\pm$0.19 & 7.8\%  \\ 
GA & 0.53$\pm$0.15 & \bf 0.99$\pm$0.54 & 92.8\% \\ \hline
\mname &  \bf 0.56$\pm$ 0.17  &  0.94$\pm$0.47  & \bf 94.0\% \\
\bottomrule[0.6pt]
\end{tabular}}
\end{minipage}
\end{table}

\noindent

\subsection*{Exp 2. Optimize Single Property}
Since most baseline models were designed to optimize single drug properties, we also conduct experiments to compare \mname with them on  optimizing the following single properties: (1) DRD; (2) QED and (3) PLogP.

From the results shown in Table~\ref{table:single}, we can see that when optimizing a single drug property, \mname still achieved the best performance overall, 
with $12.5\%$ relative higher success rate in  optimizing QED compared with the second best model GA, and  $28.8\%$ relative higher success rate in  optimizing both DRD compared with the second best algorithm VJTNN.
Among the baseline models, algorithms such as JTVAE, VJTNN, and GCPN that were designed to optimize single property  have good performance in property improvement as expected, however they generate molecules that have lower similarity hence the final success rates. Also, GA has the lowest QED and DRD improvement maybe due to its limitation in capturing global properties.
High similarity between the output and input molecules is a unique requirement for the molecule optimization task, on which \mname significantly outperformed the other baselines.

\subsection*{Exp 3. Case Study: Properties that are sensitive to local structural changes}
To further examine how \mname 
can also effectively improve properties that are sensitive to local structural change, e.g., PLogP, we show two examples in Fig.~\ref{fig:example}. For the first row, the imidazole ring in the input molecule (a) is replaced by less polar five-member rings thiazole (b and c) and thiadiazol (d). Since PLogP is related to the polarity of a molecule: more polar indicates lower PLogP. The generation results in the increase of PLogP while maintaining the molecular scaffold. For the second row, the PLogP of input molecule (e) is increased by neutralizing the ionized amine (g) or replacing with substructures with less electronegativity (f and h). These changes would also help improve the drug likeness, i.e., QED value.\\

\noindent\textbf{Sampling Efficiency}.
The sampling complexity is $O(NN_2)$ where $N$ the size of candidate set (e.g., 20) and $N_2$ is the size of the possible proposal set ($<$ 200). Empirically, this entire sampling process takes about 10-20 minutes for optimizing one source molecule, which is very acceptable for molecule optimization. And MCMC can directly operate with an unnormalized distribution which is more efficient.
Note that all the existing methods for molecule optimization also utilize RDKit in their learning process, either in preprocessing steps for creating training data~\cite{jin2018junction,jin2019learning}, or inside their training procedure such as using RDKit to evaluate reward  for reinforcement learning~\cite{You2018-xh,popova2018deep,zhou2019optimization}. 


\section{Conclusion}
\label{sec:conclusion}

In this paper, we  proposed \mname, a new MCMC sampling based method  for molecule optimization. \mname pretrains GNNs and employs three basic substructure operations  to generate new molecules and associated weights that can encode multiple  drug property constraints, upon which we accept promising molecules for next iteration. 
\mname iteratively produces new molecule candidates and can efficiently draw molecules  that satisfy all constraints. 
\mname significantly outperformed several state of the arts baselines for molecule optimization with 28.5\% to 49.1\% improvement when optimizing PLogP+QED, and PLogP+DRD, respectively.

\section*{Appendix}

\noindent\textbf{Proof of Lemma~\ref{lemma:ergodic}}. 

\begin{proof}

For a Markov chain, to guarantee its ergodicity, it is sufficient to prove its irreducibility and aperiodicity~\cite{gilks2005m}.

Regarding \textbf{irreducibility}, without loss of generalization, we need to prove that any molecule pairs ($Y$, $Z$) can communicate with each other, i.e., $Y \leftrightarrow Z$. Both Y and Z are states of the Markov chain. 

First, we want to show $Y\xrightarrow[]{} Z$, i.e., the state Y is accessible from state Z. This boils down to prove  $\exists~n\in\mathbb{N}$ such that $P^{n}_{Y,Z} > 0$. 
To show this, we construct such a Markov chain  $\{Y_0, Y_1, \cdots, Y_n\}$, where $Y_0 = Y$ and $Y_n = Z$.

First, we apply ``delete'' operation $n_1$ times to delete substructures of $Y$ until only one substructure is left, denoted C. 
That is, we have 
$Y_i \sim S_{\text{delete}} (Y_i | Y_{i-1})$ for $i = 1,\cdots, n_1$. 
Based on acceptance rate defined in Eq.~\eqref{eqn:delete_AC}, we have 
$P^{1}_{Y_{i-1}, Y_{i}} > 0$ for $i = 1,\cdots, n_1$.
$Y_0 = Y, Y_{n_1} = C$. We know $\exists~n_1\in\mathbb{N}$ such that 
\[
P^{n_1}_{Y,C} = P^{n_1}_{Y_0, Y_{n_1}} = \prod_{i=1}^{n_1} P^{1}_{Y_{i-1},Y_{i}} > 0. 
\]
Given this, we replace the substructure with a single substructure in $Z$, denoted as $C'$, 
$Y_{n_1+1} \sim S_{\text{replace}}(Y_{n_1+1} | Y_{n_1}).$
Then starting with $C' = Y_{n_1+1}$, we apply the ``add'' operation $n_2$ times until we have $Z$, 
then $
Y_{i} \sim S_{\text{add}} (Y_i | Y_{i-1})$ for $i = n_1+2, \cdots, n_1+n_2+1.$ 
where $Y_{n_1+1}=C', Y_{n_1+n_2+1} = Z$. 
Based on Eq.~\eqref{eqn:add_AC}, we have 
$P^{1}_{Y_{i-1}, Y_{i}} > 0$ for $i = n_1+2, \cdots, n_1+n_2+1.$
 $\exists~n_2\in\mathbb{N}$ such that 
\[
P^{n_2}_{C',Z} = P^{n_2}_{Y_{n_1+1}, Y_{n_1+n_2+1}} = \prod_{i=n_1+2}^{n_1+n_2+1} P^{1}_{Y_{i-1},Y_{i}} > 0. 
\] 
Thus, we have  $n=(n_1+n_2)\in\mathbb{N}$ s.t.
$
P^{n}_{Y,Z} = P^{n_1+n_2+1}_{Y,Z} \geq P^{n_1}_{Y,C} \cdot P^1_{C,C'} \cdot P^{n_2}_{C,Z} > 0.
$

Similarly, we can show  $Z\xrightarrow[]{} Y$, i.e., $\exists~n\in\mathbb{N}$ for $P^{n}_{Z,Y} > 0$. 

Now we have proved $Z \leftrightarrow Y$ hold for any molecule pairs $(Y,Z)$. Thus, we have proved irreducibility.

Next, for \textbf{aperiodicity}, there is a simple test: in Markov chain if there is a state $Y$ for which the 1-step transition probability $p(Y,Y) > 0$, then the chain is aperiodic~\cite{gilks2005m}. 
In \mname scenario, the substructure type prediction is defined in Eq.~\eqref{eqn:mask_prediction}, since it's softmax output, so for each possible substructure the probability is bounded away from zero and one. The topology prediction is defined in Eq.~\eqref{eqn:bGNN}, since it's sigmoid output, the probability is also bounded away from zero and one. 
Thus, there exists such a molecule whose the acceptance probability is lower than 1, i.e., possible to reject the proposal,  that is the 1-step transition probability is greater than 0, so aperiodicity satisfies.

\end{proof}

\noindent\textbf{Proof of Lemma~\ref{lemma:detailed_balance}}. 
\begin{proof}
In MCMC, the detailed balance condition guarantees that 
$
p(x) \calT(x\xrightarrow{} y) = p(y) \calT(y \xrightarrow{} x), 
$
where $p(\cdot)$ is the target distribution for drawing samples, $\calT(\cdot\xrightarrow{}\cdot)$ is the transition kernel from one state to another. 

Below we first show for all three proposals detailed balance condition holds for any neighboring samples ($Y_{i-1}$ and $Y_i$), then we strengthen this conclusion on the whole Markov chain.

 For ``\textbf{replace}'' proposal, we have 
\begin{equation}
\begin{aligned}
&  p_X(Y) \calT(Y \xrightarrow{} Y') \\
= & p_X(Y) \cdot [\mathrm{mGNN}(Y,v)]_{s_v} \cdot \min\{1, w_r\}  \\
 = & p_X(Y) \cdot  [\mathrm{mGNN}(Y,v)]_{s_v} 
 \\ & 
\cdot \min\bigg\{1, \frac{ p_{X}(Y') \cdot   [\mathrm{mGNN}(Y,v)]_{s'_v} }{ p_{X}(Y)\cdot  [\mathrm{mGNN}(Y,v)]_{s_v} }\bigg\} \\
= & \min\bigg\{ p_X(Y) \cdot  [\mathrm{mGNN}(Y,v)]_{s_v}, 
\\& 
\ \ \ \ p_{X}(Y') \cdot   [\mathrm{mGNN}(Y,v)]_{s'_v} \bigg\}.
\end{aligned}
\end{equation}
where we focus on replace operation on node $v$. At node $v$, $s_v$ and $s'_v$ are the actual and predicted substructure labels, respectively. 
$w_r$ is defined in Eq.~\eqref{eqn:replace_AC}.  
\begin{equation}
\label{eqn:replace_AC2}
\begin{aligned}
 w_r = \frac{ p_{X}(Y')\cdot [\text{mGNN}(Y,v)]_{s'_v} }{ p_{X}(Y)\cdot [\text{mGNN}(Y,v)]_{s_v} },
\end{aligned}
\end{equation}

For the other direction, the following equation that shows detailed balance hold for ``replace'' proposal.
\begin{equation}
\begin{aligned}
&  p_X(Y') \calT(Y' \xrightarrow{} Y) \\
= & p_X(Y') \cdot [\mathrm{mGNN}(Y,v)]_{s'_v} \cdot \min\{1, w'_r\}  \\
 = & p_X(Y') \cdot  [\mathrm{mGNN}(Y,v)]_{s'_v} 
 \\ & 
\cdot \min\bigg\{1, \frac{ p_{X}(Y) \cdot   [\mathrm{mGNN}(Y,v)]_{s_v} }{ p_{X}(Y')\cdot  [\mathrm{mGNN}(Y,v)]_{s'_v} }\bigg\} \\
= & \min\bigg\{ p_X(Y') \cdot  [\mathrm{mGNN}(Y,v)]_{s'_v}, 
\\& 
\ \ \ \ p_{X}(Y) \cdot   [\mathrm{mGNN}(Y',v)]_{s_v} \bigg\}.  \\ 
= & p_X(Y) \calT(Y \xrightarrow{} Y') \\
\end{aligned}
\end{equation}
Note that based on definition of mGNN in Eq.~\eqref{eqn:mask_prediction}, during replace operation, we have
\[
\mathrm{mGNN}(Y,v) = \mathrm{mGNN}(Y',v)
\]
We write $\mathrm{mGNN}(Y,v)$ instead of $\mathrm{mGNN}(Y',v)$ for simplicity. 

 $w'_r$ is the acceptance ratio from $Y'$ to $Y$ and satisfy 
\[
w'_r = \frac{ p_{X}(Y)\cdot [\text{mGNN}(Y',v)]_{s_v} }{ p_{X}(Y')\cdot [\text{mGNN}(Y',v)]_{s'_v} } = \frac{ p_{X}(Y)\cdot [\text{mGNN}(Y,v)]_{s_v} }{ p_{X}(Y')\cdot [\text{mGNN}(Y,v)]_{s'_v} },
\]

 For ``\textbf{add}'' proposal, we have 
\begin{equation}
\begin{aligned}
 &\  p_X(Y) \cdot \calT(Y \xrightarrow{} Y')\\
= & \ p_X(Y) \cdot  (1 - \text{bGNN}(Y, u)) ) \min\{1,w_a\} \\
= &\  p_X(Y) \cdot (1 - \text{bGNN}(Y, u) ) 
 \\ & 
 \ \ \cdot \min\bigg\{ 1,  \frac{p_X(Y') \cdot \text{bGNN}(Y, u) \cdot [\text{mGNN}(Y',v)]_{s_v} }{ p_X(Y) \cdot (1 - \text{bGNN}(Y, u)) }  \bigg\} \\
= &\  \min\bigg\{  p_X(Y) \cdot  (1 - \text{bGNN}(Y, u) ), 
\\ & 
\ \ \ \ \ \ \ \ \ \ p_X(Y') \cdot \text{bGNN}(Y, u) \cdot [\text{mGNN}(Y',v)]_{s_v}  \bigg\}
\end{aligned}
\end{equation} 
where $w_a$ is defined in Eq.~\eqref{eqn:add_AC}. 
\begin{equation}
\label{eqn:add_AC2}
\begin{aligned}
w_a = \frac{p_X(Y') \cdot \text{bGNN}(Y, u) \cdot [\text{mGNN}(Y',v)]_{s_v}  }{ p_X(Y) \cdot (1 - \text{bGNN}(Y, u)) },
\end{aligned}
\end{equation}

For the other direction, the equations below show detailed balance condition holds for ``add'' proposal. 
\begin{equation}
\begin{aligned}
 &\  p_X(Y') \cdot \calT(Y' \xrightarrow{} Y) \\
= & \ p_X(Y') \cdot 
\text{bGNN}(Y, u) \cdot [\text{mGNN}(Y',v)]_{s_v}
\cdot \min\{1,w'_a\} \\
= &\  p_X(Y') \cdot 
\text{bGNN}(Y, u) \cdot [\text{mGNN}(Y',v)]_{s_v}
 \\ & 
 \cdot \min\bigg\{ 1,  \frac{ p_X(Y) \cdot  (1 - \text{bGNN}(Y, u)) }{ p_X(Y') \cdot  \text{bGNN}(Y, u) \cdot [\text{mGNN}(Y',v)]_{s_v} } \bigg\} \\
= &\  \min\bigg\{ \  p_X(Y') \cdot \text{bGNN}(Y, u) \cdot [\text{mGNN}(Y',v)]_{s_v}, 
\\ & 
\ \ \ \ \ \ \ \ \ \ \ \ \ p_X(Y) \cdot (1 - \text{bGNN}(Y, u)) \bigg\}
\end{aligned}
\end{equation} 
where $w'_a$ is the acceptance rate from $Y'$ to $Y$ using delete operation, defined in Equation~\eqref{eqn:delete_AC}. 

For the  ``\textbf{delete}'' proposal, we can view it as the reverse procedure of the ``add'' proposal, which is easy to prove.

Since detailed balance holds for all proposals, i.e., 
$
  p_X(Y_i) \calT(Y_i \xrightarrow{} Y_{i-1}) = p_X(Y_{i-1}) \calT(Y_{i-1} \xrightarrow{} Y_{i}),
$ by integrating out $Y_{i-1}$ on both sides, we obtain 
\begin{equation}
\begin{aligned}
 p_X(Y_i)
= &  \int p_X(Y_i) \calT(Y_i \xrightarrow{} Y_{i-1}) dY_{i-1} \\
= &  \int p_X(Y_{i-1}) \calT(Y_{i-1} \xrightarrow{} Y_{i}) dY_{i-1} \\ 
\end{aligned}
\end{equation}
That is, distribution $p_X(\cdot)$ is stationary with a transition kernel $\calT(\cdot\xrightarrow{}\cdot)$, i.e., 
$
\calT(p_X) = p_X 
$
hold for transition kernel that contains ``replace'', ``add'' or ``delete'' proposal. 

\end{proof}

\section{More Experimental Details}

\noindent
\textbf{Implementation} We implemented \mname using Pytorch 1.0.1 and Python 3.7 on an Intel Xeon E5-2690 machine with 256G RAM and 8 NVIDIA Pascal Titan X GPUs. 
 We use Adam optimizer with a learning rate of 0.001. For pretraining, we follow~\cite{hu2019strategies} to set GNNs with 5 layers and 300-d hidden units. For fully connected layer used in Equation~\eqref{eqn:softmax} and ~\eqref{eqn:bGNN2}, we use two layer feedforward NN, the hidden size is 50. 
Tanimoto similarity, PLogP, and QED scores were computed using RDKit package~\cite{landrum2006rdkit}. 
The DRD2 activity prediction model is publicly available~\footnote{\url{https://github.com/MarcusOlivecrona/REINVENT}}. 
For mGNN, we randomly mask a single node for each molecule, while for bGNN, we randomly select a leaf node for each molecule. 
When training them, we choose batch size 256, epochs number 10, and learning rate $1e^{-3}$, 
Then, during inference stage, we keep 20 molecules each iteration, i.e., in Algorithm~\ref{alg:main}, $N = 20$. We set $T_{\text{max}} = 10$ and $T_{\text{burnin}} = 5$. 
When optimizing ``QED+PLogP'' and ``DRD+PLogP'' ($M=2$), for the target distribution $p_X(Y)$ (defined in Eq.~\eqref{eqn:target_distribution}), we have $\eta_0 = 1.0$, $\eta_1 = 0.3$, and $\eta_2 = 0.3$.

\subsection*{Node and Edge Feature}
In this paper, a substructure corresponds to a node, which contains 149 different types, including 118 atoms (e.g., Carbon, Nitrogen, Oxygen, Sulfur, etc) and 31 frequent single rings (e.g., Benzene, Cyclopropane). 
Each edge corresponds to a bond in molecular graph. There are totally 4 kinds of bonds, thus we have 4 edge type in total, including single, double, triple and aromatic.

\noindent\textbf{Success Rate}
 For LogP, we define a success as $\text{sim}(X,Y) \geq 0.4$ and $\text{PLogP}(Y) - \text{PLogP}(X) \geq 0.5$. For QED, we define a success as $\text{sim}(X,Y) \geq 0.4$ and $\text{QED}(Y) - \text{QED}(X) \geq 0.1$. For DRD, we define a success as $\text{sim}(X,Y) \geq 0.4$ and $\text{DRD}(Y) - \text{DRD}(X) \geq 0.2$. For  optimizing both QED and PLogP , we define a success as (i) $\text{sim}(X,Y) \geq 0.3$, (ii) $\text{QED}(Y) - \text{QED}(X) \geq 0.1 $, (iii) $\text{PLogP}(Y) - \text{PLogP}(X) \geq 0.3$. For optimizing both DRD and PLogP, we define a success as (i) $\text{sim}(X,Y) \geq 0.3$, (ii) $\text{DRD}(Y) - \text{DRD}(X) \geq 0.2$, (iii) $\text{PLogP}(Y) - \text{PLogP}(X) \geq 0.3$. \\

\noindent\textbf{Baseline Setup} Below are more detailed about baselines. \\

\noindent
\textbf{JTVAE}. We follow~\cite{jin2018junction} to use 780 substructures for junction tree, 5-layers message passing network  as encoder. The hidden state dimension is set to 450. For the graph encoder, the initial atom features include its atom type, degree, its formal charge and its chiral configuration. Bond feature is
a concatenation of its bond type, whether the bond is in a ring, and its cis-trans configuration. 
For the tree encoder, each cluster is represented with a neural embedding vector. The tree and graph decoder use the same feature setting as encoders. The graph encoder and decoder run three iterations of neural message passing.

\noindent
\textbf{VJTNN}. We follow ~\cite{jin2019learning} to set the hidden state dimension as 300 and latent code dimension as 8. 
The tree encoder runs message passing for 6 iterations, and graph encoder runs for 3 iterations. The hidden state dimension of the recurrent encoder and decoder is set to 600. 
We leverage the Adam optimizer for 20 epochs with learning rate 0.001. The learning rate is annealed by 0.9 for every epoch. For adversarial training, the discriminator has 3-layer and with hidden layer dimension 300 and LeakyReLU activation function. 

\noindent
\textbf{GCPN}. We follow ~\cite{You2018-xh} to set up an OpenAI Gym environment using RDKit package. 
The maximum atom number is 38. Since we represent molecules in kekulized form, there are 9 atom types and 3 edge types. 
We use a 3-layer GCPN as the policy network with 64 dimensional node embedding in all hidden layers. Batch normalization is applied after each layer.
We use Adam optimizer with batch size 32. The learning rate is set as 0.001. 

\noindent
\textbf{GA}.  we follow~\cite{Nigam2020Augmenting} to run each experiment for 20 generations with a population size of 500. 
We compute the number of experiments that successfully proposed molecules with a squared difference less than 1.0. Each run is constrained to run for 100 generations with a maximum canonical SMILES length of 81 characters.



%

\bibliographystyle{plain}
\bibliography{ref}




\end{document}